\documentclass{article}

\usepackage[accepted]{icml2025}

\usepackage[colorlinks=true,allcolors=perfblue]{hyperref}
\usepackage{graphicx}
\usepackage{empheq}
\usepackage[utf8]{inputenc}
\usepackage[T1]{fontenc}
\usepackage{url}
\usepackage{booktabs}
\usepackage{amsfonts}
\usepackage{amsmath}
\usepackage{amssymb}
\usepackage{nicefrac}
\usepackage{microtype}
\usepackage{natbib}
\usepackage{doi}
\usepackage{algorithm}
\usepackage{makecell}
\usepackage{enumitem}
\usepackage{caption}
\usepackage{subcaption}
\usepackage{pifont}
\usepackage{tikz}
\usetikzlibrary{arrows,calc,matrix,backgrounds}
\newcommand{\tikzAngleOfLine}{\tikz@AngleOfLine}
\def\tikz@AngleOfLine(#1)(#2)#3{%
\pgfmathanglebetweenpoints{%
\pgfpointanchor{#1}{center}}{%
\pgfpointanchor{#2}{center}}
\pgfmathsetmacro{#3}{\pgfmathresult}%
}
\usepackage{tikz-cd}
\usepackage[most]{tcolorbox}
\usepackage{footmisc}
\usepackage{xcolor}
\usepackage{amsthm}
\usepackage{thmtools}
\usepackage{float}
\usepackage{bm}

\usepackage{dsfont}
\usepackage[customcolors]{hf-tikz}

\definecolor{expert}{HTML}{008000}
\definecolor{error}{HTML}{f96565}
\definecolor{learner}{HTML}{F79646}
\definecolor{perfblue}{RGB}{64, 114, 175}

\theoremstyle{plain}

\theoremstyle{definition}

\theoremstyle{remark}

\declaretheoremstyle[
headfont=\normalfont\itshape,
qed=\qedsymbol,
]{mypf}

\renewcommand{\algorithmiccomment}[1]{// #1}

\usepackage{nicematrix}

\icmltitlerunning{Pretrained Joint Predictions for Scalable Batched Bayesian Optimization}

\begin{document}

\twocolumn[
\icmltitle{Pretrained Joint Predictions for Scalable Batch Bayesian Optimization of Molecular Designs}

\begin{icmlauthorlist}
\icmlauthor{Miles Wang-Henderson}{}
\icmlauthor{Benjamin Kaufman}{}
\icmlauthor{Edward Williams}{}
\icmlauthor{Ryan Pederson}{}
\icmlauthor{Matteo Rossi}{}
\icmlauthor{Owen Howell}{}
\icmlauthor{Carl Underkoffler}{}
\icmlauthor{Narbe Mardirossian}{}
\icmlauthor{John Parkhill}{}
\end{icmlauthorlist}

\begin{center}
Terray Therapeutics
\end{center}

\vskip 0.3in
]

\begin{abstract}
Batched synthesis and testing of molecular designs is the key bottleneck of drug development. There has been great interest in leveraging biomolecular foundation models as surrogates to accelerate this process. In this work, we show how to obtain scalable probabilistic surrogates of binding affinity for use in Batch Bayesian Optimization (Batch BO). This demands parallel acquisition functions that hedge between designs and the ability to rapidly sample from a joint predictive density to approximate them. Through the framework of Epistemic Neural Networks (ENNs), we obtain scalable joint predictive distributions of binding affinity on top of representations taken from large structure-informed models. Key to this work is an investigation into the importance of prior networks in ENNs and how to pretrain them on synthetic data to improve downstream performance in Batch BO. Their utility is demonstrated by rediscovering known potent EGFR inhibitors on a semi-synthetic benchmark in up to 5x fewer iterations, as well as potent inhibitors from a real-world small-molecule library in up to 10x fewer iterations, offering a promising solution for large-scale drug discovery applications.
\end{abstract}

\section{Introduction}
\label{sec:intro}

Small-molecule discovery campaigns necessitate the parallel and sequential selection of compounds in repeated rounds of Design-Make-Test-Analyze (DMTA) cycles, to address various molecular properties such as binding affinity. This process is naturally framed as Batch Bayesian Optimization (Batch BO) \cite{garnett_bayesoptbook_2023}. Success demands two critical desiderata: (1) parallel acquisition functions that hedge between selected designs, and (2) joint predictive distributions that capture the correlations necessary for such acquisition functions \cite{wen2021predictions}.

In practice, to employ parallel acquisition functions, one requires the ability to efficiently sample jointly from a probabilistic surrogate $p_\theta(y_{1:N} | x_{1:N})$ in order to capture the correlations between candidate designs. Gaussian processes (GPs) are the canonical probabilistic surrogate used for Batch BO for this reason, as they can provide exact \emph{joint} posterior inference, straightforwardly accessible via sample paths. However, it is well known that their cubic scaling makes them cumbersome for conditioning on the large datasets used for binding affinity regression, and several approximations exist to remedy this issue. Other methods that use neural networks for constructing probabilistic surrogates, such as deep ensembles or ensemble-based Bayesian Neural Networks (BNNs) \cite{lakshminarayanan2017deep, arbel2023primerBNN, duffield2024scalable}, can allow principled conditioning to be amortized to training and be used to perform inference via Monte Carlo sampling. However, few focus on the design of good joint predictive distributions that can be sampled from rapidly \cite{wen2021predictions}.

\setlength{\textfloatsep}{1pt}
\begin{figure}[h]
  \centering
  \vspace{0.2em}
  \includegraphics[width=0.5\textwidth]
  {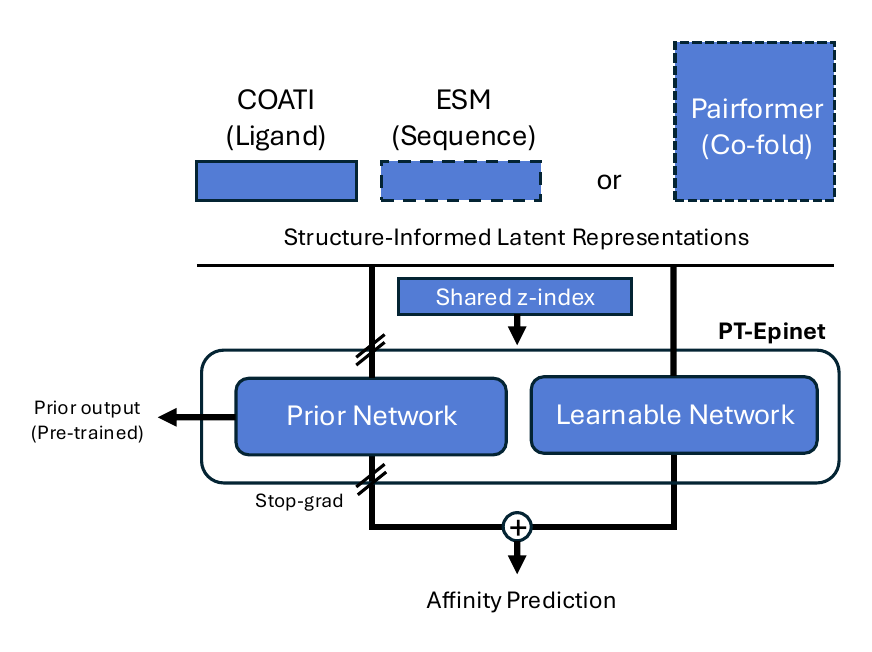}
   \caption{Simplified overview of an ENN-based architecture that uses a pretrained prior network, with optional inclusion of different fixed latent representations as input. In our experiments, we use COATI, a ligand-only representation \cite{kaufman2024coati}, to address a single target.}
  \label{fig:priorbind_summary}
\end{figure}

A solution to this problem is particularly important, as recent work has shown that using latent representations from foundation models of structural biology, e.g., co-folding models and those trained on large amounts of chemistry-specific data, can be used to build predictive models that approach the accuracy of physics-based methods  \cite{passaro2025boltz2}.

Our work investigates how to address these design desiderata through the framework of Epistemic Neural Networks (ENNs) \cite{osband2021epistemic}, enhanced with pretrained prior functions. Our main contributions are as follows:
\begin{itemize}[leftmargin=2em]
    \item Pretrained prior functions in ENNs and a comparison of their resulting joint predictive distributions with those of hand-designed random prior architectures. We show that pretraining the prior network using a reference process yields improved joint log-loss which translates to better Batch BO performance.
    \item Relevant real-world benchmarks demonstrating the scalability of joint samples from pretrained ENNs that use latent-representations from a large foundation model of chemistry \cite{kaufman2024coati}. We assess the ability of ENNs to drive sample-efficient parallel acquisition of potent small-molecules on two realistic binding affinity optimization tasks. The first is re-discovering EGFR inhibitors from a public dataset \cite{liu2025bindingdb}, and the second is re-discovering the most potent small-molecule binders from a large proprietary dataset for a pipeline target at Terray.
\end{itemize}

\section{Preliminaries}
\label{sec:prelim}

\subsection{Batch Bayesian Optimization}

Molecular property optimization campaigns are often performed by successively synthesizing and assaying designs in parallel. This is done for several practical reasons such as efficiencies of scale and low-odds for success for any individual design.

To optimally select designs simultaneously requires us to consider how their properties are correlated in order to hedge between them. Thus, the primary challenge is obtaining \emph{joint} predictive distributions $p_\theta\left(y_{1:N} | x_{1:N}\right)$ via some surrogate model, capturing the correlation structure essential for balancing exploitation of properties we want to maximize with the hedging of selected designs within a batch. Any selection strategy for this problem can be formulated as finding a batch $B$ of designs with the highest expected utility under a joint predictive distribution: $$\mathbb{E}_{\hat{y}_{1:N} \sim p(y_{1:N})}\left[u(\hat{y}_{1:B})\right]$$ While such expectations are analytically intractable and difficult to optimize, they admit Monte-Carlo approximations using functional particles from a probabilistic surrogate, e.g., GP sample paths \cite{wilson2017reparameterization, garnett_bayesoptbook_2023}. For simplicity, we focus on two complementary parallel acquisition functions:

\textbf{qPO}: $\mathbb{E}_{p(y_{1:N})}\left[\max(\hat{y}_{1:N}) \leq \max(\hat{y}_{1:B})\right]$. The probability of a batch containing the global maximum \cite{hennig2011entropy, fromer2024batched, menet2025lite}. By integrating over a joint density, qPO penalizes correlations between designs and hedges within an acquired batch. qPO has empirically shown to be useful on real examples of Batch BO \cite{fromer2024batched}, though care must be taken in that its approximation requires many iid sample paths to converge, and scales linearly with size of the pool due to the use of joint sample paths $\hat{y}_{1:N} = (\hat{y}_1, \ldots, \hat{y}_N)$ \cite{menet2025lite}.

\textbf{EMAX}: $\mathbb{E}_{p(y_{1:B})}\left[\max(\hat{y}_{1:B})\right]$. Expected maximum value in batch \cite{azimi2010batch}, sometimes called parallel simple-regret (qSR) \cite{wilson2017reparameterization}. EMAX also penalizes correlations between designs in a batch. We forgo greedy optimization since EMAX is not submodular \cite{azimi2010batch}, i.e., building up the batch constructively by adding designs. An advantage of EMAX over qPO is that its evaluation cost depends only on the batch size, not the pool size $N$. We use a simple stochastic local-search procedure that uses single random swaps of the batch to optimize the inner-loop.

\begin{figure}[h]
  \centering
  \includegraphics[width=0.4\textwidth]{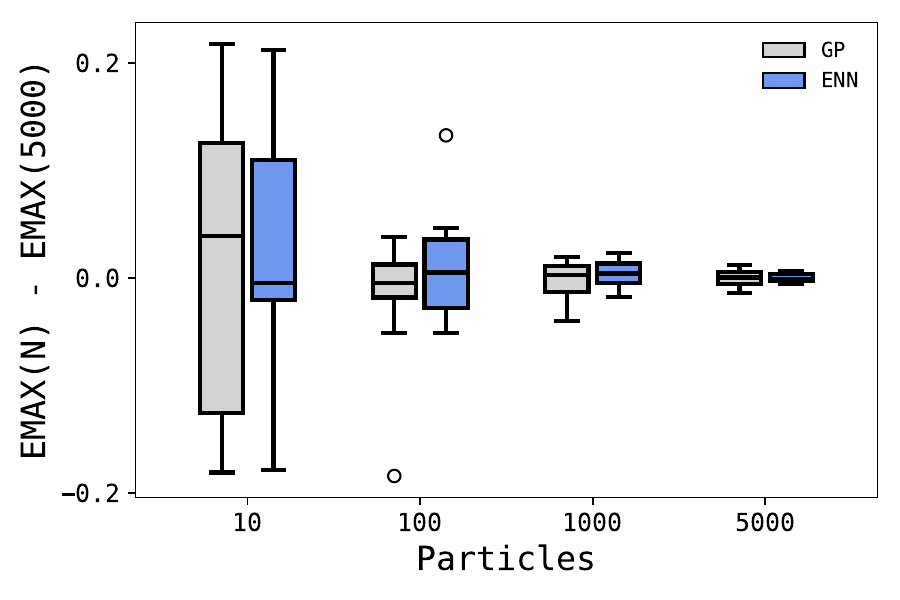}
  \caption{A scalable strategy for sampling is necessary to yield convergent estimates for Batch BO. Estimates of the expected maximum (EMAX) pIC50 of a batch of 25 compounds as a function of number of particles under 10 draws, e.g., sample paths from a Gaussian Process (GP) or epistemic index draws from an Epistemic Neural Network (ENN). This shows error in estimate is negligible after the number of particles is proportional to square of batch size.}
  \label{fig:particle_convergence}
  \vspace{.5em}
\end{figure}

Empirically, selecting $B$ designs from a set of $N$ means employing global strategies like qPO can require $K \in \Theta(N^2)$ particles (sample paths) to obtain convergent estimates \cite{menet2025lite}, and $\Theta(B^2)$ for local strategies like EMAX, which may be more practical in screening settings where the pool size is very large. See Figure \ref{fig:particle_convergence} for estimates of the expected maximum pIC50 of a small batch of compounds as we increase the number of samples.

\subsection{Ensembles for Approximate Inference}

A simple method to construct probabilistic surrogates for large datasets is via deep ensembles \citep{lakshminarayanan2017deep, wild2023rigorous}, in which copies of the same network architecture with randomly sampled parameter initializations $\{\theta_{1}^{(0)}, \ldots, \theta_{K}^{(0)}\} \sim p(\theta)$ are independently trained to maximize data likelihood. Due to stochasticity in the optimization routine and the general non-convexity of the loss function, randomly initialized particles usually converge to different local minima, yielding the following approximation to a posterior that marginalizes over parameters:

$$p(\theta | \mathcal{D}) \approx \frac{1}{K}\sum_{k=1}^{K} \delta\left(\theta_{k}^{(t)} = \theta \right)$$

Where $\delta(x = \cdot)$ is the Dirac delta centered at $x$. This can then be converted to a posterior predictive that can also be used to approximate epistemic entropy:

$$p(y | x, \mathcal{D}) \approx \frac{1}{K}\sum^K_{k=1} \delta\left(f_{\theta_k^{(t)}}(x) = y\right)$$

This straightforward weight-space perspective has been experimentally validated and theoretically motivated through several works \cite{floge2024stein, loaiza2025deep, duffield2024scalable}. Despite their good empirical performance, the need to retain copies of all network parameters (or subsets of parameters) makes this approach cumbersome to employ in Batch BO acquisition functions, as MC approximations have errors that decay slowly at a rate of $K^{-1/2}$ particles.

\subsection{Epistemic Neural Networks}

Epistemic Neural Networks (ENNs) \cite{osband2021epistemic} are a similar ensemble method that can be used to obtain joint predictive distributions by instead marginalizing over a latent distribution $p_Z(z)$ of particles or "epistemic indices", in the language of ENNs. For example, in the case of regression:
$$p(y_{1:N}|x_{1:N}) = \int_z \delta\left([f_\theta(x_i, z)]_{1:N} = y_{1:N}\right) \ p_Z(z) \ \mathrm{d}z$$

This function-space approach is an important feature of ENNs, as marginalization is not performed over parameters $\theta$, but rather over a latent index $z$. Thus, depending on the definition $p_Z(z)$ and how the network is designed to condition on the latent variable $z$, ENNs can remain parameter-efficient and model a joint predictive distribution cheaply.

The regression objective of ENNs is simply to minimize the marginal regularized squared loss, i.e., for a batch size of one:
$$\min_\theta \mathcal{L}(\theta, x, y, z) = (f_\theta(x, z) - y)^2 + \lambda R(\theta)$$

Given this marginal loss only encourages ENNs to reduce epistemic entropy near training data, a key design component of ENNs is the use of additive prior functions, which are frozen networks $f_\phi(x, z)$ conditioned on the same latent variable $z$ \cite{dwaracherla2022ensembles}. These greatly improve the \emph{joint} predictive distribution after training \cite{osband2021epistemic}. 

Intuitively, good initial specification of its architecture, or pretraining of its parameters, induces a meaningful prior distribution over functions $p(f)$ \cite{tran2020goodfunctionalprior}. And through regularization of the trainable parameters of the ENN or its predictions, one can tune divergence from this prior, e.g., via L2 from the parameters at initialization $R(\theta) = \|\theta - \theta_0\|_2$ \cite{dwaracherla2022ensembles}.

\subsection{Functional Priors}

Instead of explicitly encoding inductive biases about the function in the construction of a neural network's architecture, or implicitly in its hyperparameters, our approach aims to absorb this prior knowledge in a data-driven manner from a synthetic reference process \cite{tran2020goodfunctionalprior, rothfuss2021metalearning, fan2022hyperbo, cinquin2025what, treven2025simulation}.

For example, it has been noted empirically that ensemble-based Bayesian Neural Networks (BNNs) with weight-space priors are particularly sensitive to hyperparameters \cite{cinquin2025what, arbel2023primerBNN}, and there is ambiguity as to what functions they induce. Regularizing directly in function-space, however, allows for a more direct application of the user's beliefs. And it has been shown in several works that regularizing neural network surrogates with a GP prior is more robust in real-world BO \cite{rothfuss2021metalearning, cinquin2025what}. 

The same function-space regularization can be accomplished here via the additive prior functions used in ENNs by pretraining a prior network $f_\phi(x, z): \mathcal{X} \times \mathcal{Z} \rightarrow \mathbb{R}$ to reproduce a sample path of a reference stochastic process $h(x, z): \mathcal{X} \times \mathcal{Z} \rightarrow \mathbb{R}$ with epistemic index $z \sim p_Z$. More formally, for any finite collection of data $(x_1, \ldots, x_N)$ we aim to estimate a realization of an infinite-dimensional random vector $(h(x_1, z), \ldots, h(x_N, z))$ indexed by an epistemic particle. Afterwards, during typical ENN training, the prior network remains fixed and divergence from it is tuned by regularizing the weights of the learnable network.

In this work, we accomplish this pretraining on synthetic datasets of GP sample path realizations. \citeauthor{treven2025simulation} has similarly used a physics-based simulator as a reference process, where each realization is indexed via its hyperparameters. Others have approached matching distributions over functions analogously, e.g. Neural Diffusion Processes \cite{dutordoir2023neural}. Inspired by the success of these approaches, and to reduce iterative sampling costs, in our example experiments we use a highly simplified one-step squared loss.

$$\min_\phi \mathcal{L}(\phi) = \mathbb{E}_{z \sim p_Z}\mathbb{E}_{x \sim \mathcal{D}}\left[\big(f_\phi(x, z) - h(x, z)\big)^2\right]$$

\section{Pretrained Priors for Binding Affinity}
\label{sec:method}

In this section we discuss details of our approach, and provide pedagogical examples on synthetic datasets.

\subsection{The Epinet Architecture}

The canonical architecture for ENNs as introduced by \citealp{osband2021epistemic} is the Epinet:
$$f_\text{ENN}(x, z) = \mu_\zeta(x) + f_\eta(\text{sg}[\tilde{x}], z) + f_\phi(\text{sg}[\tilde{x}], z)$$

Where $\theta=(\zeta, \eta)$ are trainable parameters,  $\mu_\zeta$ acts as a point-estimate of the function mean, $f_\eta$ is a trainable component that learns a data-driven correction to the prior, and $f_\phi$ is a frozen additive prior function with randomly initialized parameters $\phi$. It is possible for this network to use detached (stop-grad) $\text{sg}[ \cdot ]$ hidden activations $\tilde{x}$ from a pretrained base network with frozen parameters, e.g., the last-layer of $\mu_\zeta$ or hidden representations from a structure-informed foundation model \cite{kaufman2024coati, Abramson:2024vm, passaro2025boltz2}, allowing one to use the Epinet as a cheap adapter. 

As the choice of architecture and initialization for $f_\phi$ defines the prior function, its design can critically impact the correlation structure of the Epinet's samples, and correspondingly the approximations to the acquisition functions employed in Batch BO. In the example introduced by \citeauthor{osband2021epistemic}, the prior and learnable network is simply $\text{mlp}(\cdot)^{\top}z$ and in follow-up work on active learning they further simplify the architecture to use a \emph{linear} layer \cite{osband2022finetuning}. While a trained MLP with ReLU activations can produce interesting non-linear functions, at initialization the result is an approximately piecewise affine prior on $\tilde{x}$ with Gaussian marginals. As suggested by the authors, however, careful design of the prior network can lead to better joint predictive distributions. 

This is particularly important in the context of drug discovery, where properties can be rough and difficult to model. Empirically, others have found that simple linear priors tend to under-perform, even when combined with last-layer representations from a large foundation model of chemistry \cite{cinquin2025what}.

\begin{figure}[t]
  \centering
  \includegraphics[width=0.475\textwidth]{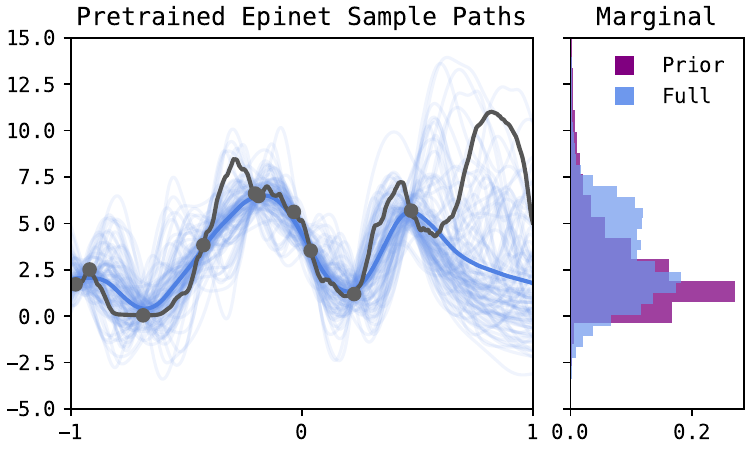}
  \caption{Left: Sample paths from the joint predictive distribution of a Pretrained Epinet with a frozen prior network, after training the learnable component on 10 observations. Paths drawn in blue using $K=100$ epistemic particles. True function and training points in grey. We see that the joint predictive distribution is well-calibrated and covers the true function. Right: empirical marginal density of samples from $f_\theta + f_\phi$ in blue, and prior network only $f_\phi$ in purple.}
  \label{fig:1d_PTepinet}
  \vspace{1em}
\end{figure}

\subsection{Pretraining the Prior Network}

Given that the design of $f_\phi$ is critical for joint predictions, we evaluate variants of the Epinet with either pretrained prior functions or inductive biases in the architecture. For example, the user could approximately encode a belief about the complexity of their objective function by embedding input features or hidden activations $\tilde{x} \in \mathbb{R}^d$ via Random Fourier Features:
$\text{RFF}(x) = \sqrt{2/d} \cos\left(Wx  + b\right)$
where $W \sim \mathcal{N}(0, \ell^{-2}\mathbb{I})$, $b \sim \mathrm{Unif}(0, 2\pi)$, and $\ell$ is a lengthscale hyperparameter \cite{rahimi2007random}. 

Rather than being explicit by hard-coding the prior only via architectural changes, one can also choose to pretrain the prior network with data, summarized in Algorithm \ref{alg:pt_epi}.

\begin{algorithm}[h]
   \caption{Pretraining the Prior Network}
   \label{alg:pt_epi}
\begin{algorithmic}
   \STATE \textbf{Input:} 
   \STATE \hspace{1em} Initial prior weights $\phi$ and distribution $p_Z(z)$
   \STATE \hspace{1em} Reference process $h(x,z)$, e.g. GP prior
   \STATE \hspace{1em} Dataset $D_N = \{x_j\}_{j=1}^N$, e.g. uniform from $[0, 1]^d$
   \STATE \hspace{1em} Particle count $K$, batch size $M$
   \STATE

   \WHILE{not converged}
      \STATE \textbf{1.} Draw $K$ particle indices $\{z_i\}_{i=1}^K \sim p_Z$, and form minibatch $\{x_j\}_{j=1}^M$ from $D_N$.
      \vspace{0.1em}
      \STATE \textbf{2.} Sample joint paths for each particle:
      \vspace{0.1em}
      \STATE \hspace{2em} $y_{i, 1:M} = \big(h(x_1, z_{i}),\,h(x_2,z_{i}),\,\dots,\,h(x_M,z_{i})\big)$
      \STATE \textbf{3.} Form minibatch and update $\phi$ to minimize
      \vspace{0.1em}
      \STATE \hspace{2em} $\displaystyle
      \sum_{ij}(f_\phi(x_j, z_i) - y_{ij})^2
      $
   \ENDWHILE

   \STATE \textbf{Return:} Trained prior $f_\phi(x,z)$.
\end{algorithmic}
\end{algorithm}

To assess this approach in a toy example, we use a warped GP as a reference process to generate synthetic datasets of samples with non-Gaussian marginals, mimicking the bounded and skewed distributions of binding affinity measurements seen in real assays. We sample latent paths $(\tilde{h}(x_1), \ldots, \tilde{h}(x_N)) \sim \mathrm{GP}(0, k_{\text{Mat\'ern}32})$ and apply an element-wise warp to obtain our final labels $h(x) = \text{warp}(\tilde{h}(x))$. We use a simple sigmoid-power $\text{warp}(\tilde{h}; a, b, c) = S((\tilde{h} - a)/b)^c$ where $S(\cdot)$ is the sigmoid function and $a, b, c$ are warping parameters that control location, scale, and skewness. This serves both as a means to generate the datasets used to pretrain the prior network, as well as test unseen sample-paths from a known and well-specified distribution over functions for evaluation.

\begin{figure}[h]
  \centering
  \includegraphics[width=0.487\textwidth]{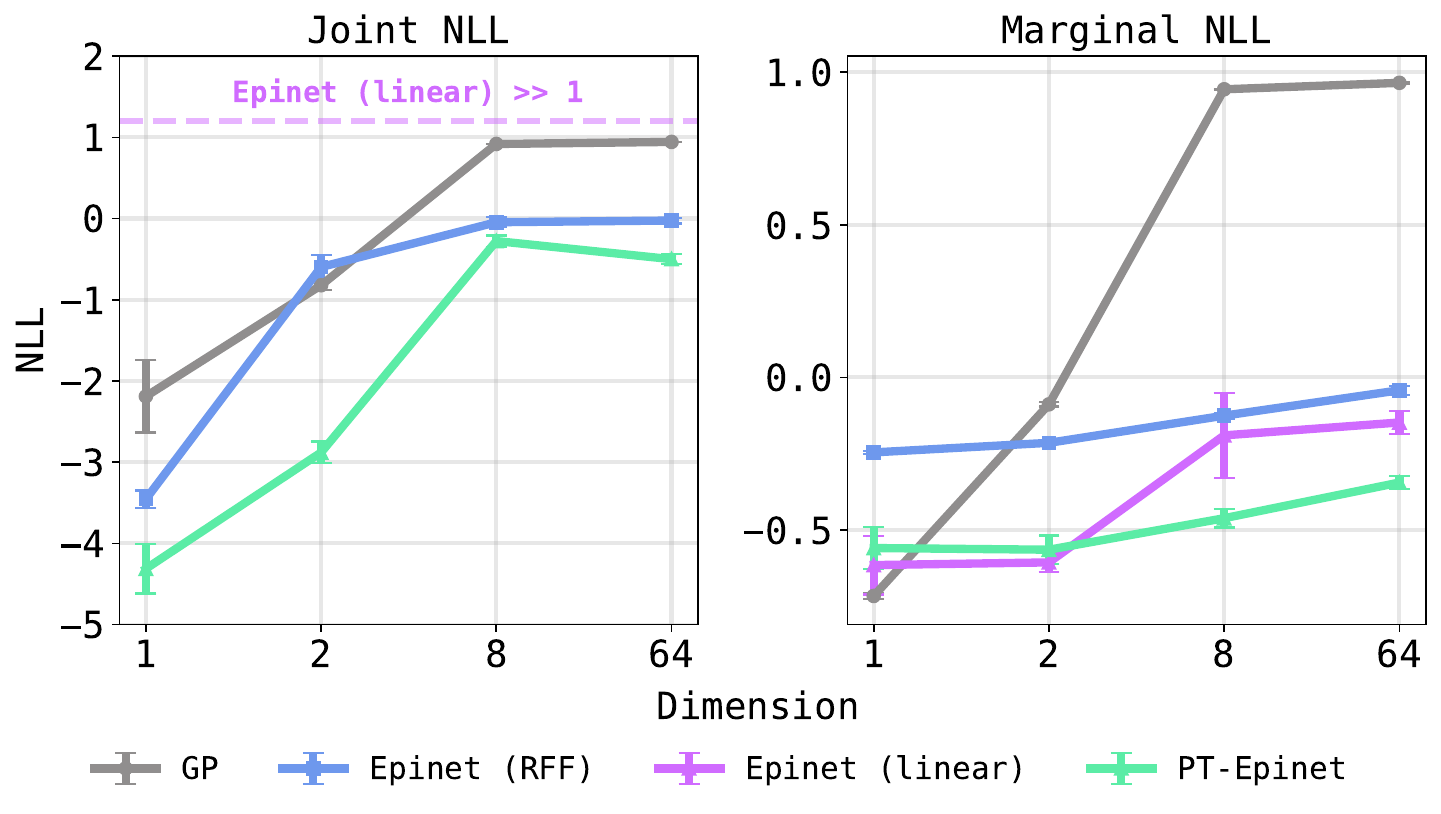}
  \caption{Comparison across different input-dimensions of negative log-loss on a test subset after training on a small subset of a warped GP sample-path. Epinet variants are not strongly distinguished on marginal negative log-loss (NLL). As expected, the Pretrained (PT) Epinet does consistently well, also on joint negative log-loss evaluated using augmented dyadic sampling \cite{osband2021epistemic}.}
  \label{fig:nll_comparison}
\end{figure}

Additionally, to encourage the Epinet to produce non-Gaussian marginals, our Pretrained Epinet architecture is further simplified by dropping the mean function provided by a base network. In Figure \ref{fig:1d_PTepinet} we can see that this variant is able to generate non-Gaussian marginals. The importance of the resulting joint predictions is illustrated in Figure \ref{fig:nll_comparison}, showing that pretrained and RFF prior functions achieve better joint negative log-loss (NLL) compared to linear ones on evaluation. 

Unseen test labels are obtained by sampling single warped GP paths using a Mat\'ern32 kernel with lengthscale $\ell \propto \sqrt{d}$ over input-dimensions $d \in [1, 2, 8, 64]$, and inputs $x$ are chosen uniformly from the unit hypercube $[0, 1]^d$. We use warping parameters $a, b, c = (0.5, 1.5, 2.0)$ to generate bounded sample-paths $h \in [0, 1]$. We also show an exact GP baseline with $\ell \propto \sqrt{d}$ \cite{hvarfner2024vanilla}.

All Epinet variants perform similarly across dimensions on marginal negative log-loss, whereas only prior functions that are well-specified yield good joint predictions, with Pretrained (PT) Epinet performing the best. This suggests that we can expect improved hedging and performance in Batch BO when parallel acquisition functions are powered by samples from a pretrained Epinet's joint predictive distribution, which we demonstrate in more extensive experiments.

\begin{figure*}[!h]
  \centering
  \includegraphics[width=0.485\textwidth]{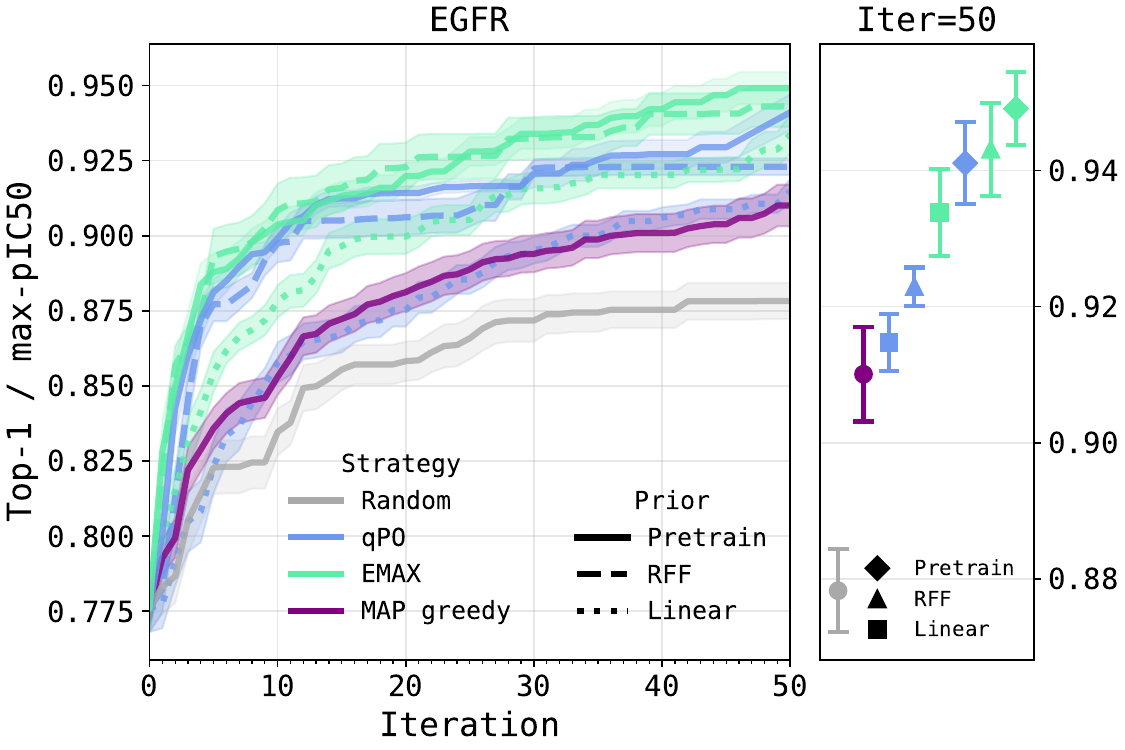}
  \hfill
   \includegraphics[width=0.485\textwidth]{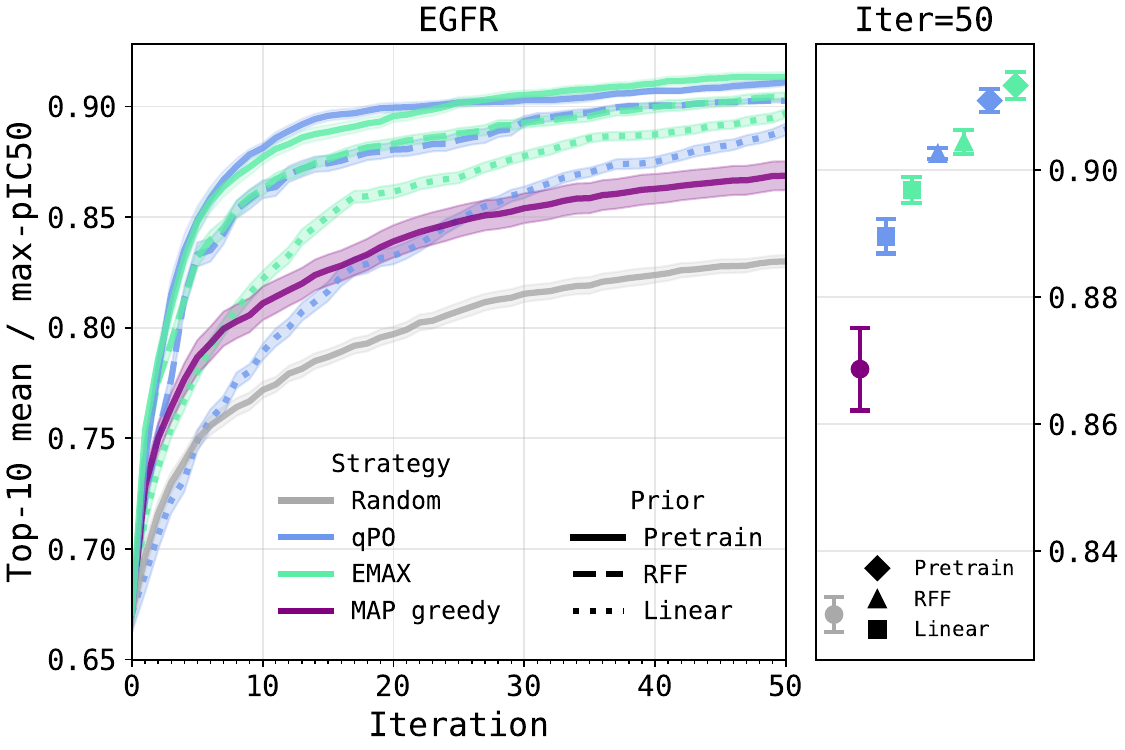}
  \caption{Performance of different Epinet variants and acquisition functions in maximizing pIC50 on the EGFR dataset. Compared to a greedy baseline, using Pretrained Epinets allow us to retrieve the same Top-1 pIC50 in 5x fewer iterations and the same Top-10 mean pIC50 in 7x fewer iterations. Moreover, the final iteration yields more potent molecules than other baselines. An absolute improvement of 0.1 in normalized Top-1 pIC50 retrieved corresponds to an approximately 14x reduction in IC50 concentration. For all curves we plot the mean and standard errors over 20 random seeds. Left: y-axis shows Top-1 pIC50 retrieved per iteration, normalized by the true maximum pIC50 in the dataset. Right: shows the mean pIC50 of the Top-10 highest retrieved compounds, also normalized. See also Table \ref{tab:egfr_results}.}
  \label{fig:egfr_scaling}
\end{figure*}

\section{Experiments and Results}
\label{sec:results}

\subsection{EGFR Inhibitor Screening}

In this experiment, we assess combinations of Epinet variants and parallel acquisition functions on a relevant real-world task, namely screening potent inhibitors of EGFR.

\textbf{Dataset and Embeddings.} We use 13,201 unique publicly reported EGFR inhibitors extracted from BindingDB \cite{liu2025bindingdb}. These contain mostly active compounds and many are highly potent. To simulate a realistic program of iterated design cycles where most compounds are inactive, and the available pool of compounds is diverse, we augment this dataset with synthetically generated and labeled compounds. We sample these "decoy" compounds using particle guidance \cite{corso2023particle} with a latent diffusion model of small molecules, COATI-LDM \cite{kaufman2024latent}, trained on more than one billion drug-like molecules. The total number of unique compounds is 46,000. We then fit a GP with a Mat\'ern32 kernel to the 768-dimensional COATI embeddings \cite{kaufman2024coati} of the public set. Using a single sample path from the conditioned GP we then label this diverse decoy subset after applying our sigmoid-warp with bounds $[0, 8]$ and $a, b, c = (6, 1, 3)$. These parameters were manually chosen such that the most potent inhibitors in the full dataset come from the public set of known inhibitors.

\textbf{Epinet Variants.} We employ an Epinet as our main surrogate with three different architectures that induce different prior functions, and use COATI embeddings as the input. Following the architecture introduced by \cite{osband2021epistemic}, the Linear and RFF variants have prior networks that remain fixed from random initialization. The RFF variant differs only in that it pre-embeds the input representations before feeding to the final layer using a length-scale of $\ell = 0.5 \sqrt{768}$.

The third variant uses a pretrained prior network, and omits the mean base network $\mu_\zeta$ to encourage asymmetric marginals. To assess the potential of this setup in a case where the functional prior is well-specified but not an oracle prior, the prior network is trained on sample paths from a warped GP prior with similar warp hyperparameters as used to label the decoy partition of the EGFR dataset $a,b,c=(0.1, 0.75, 2.0)$; however, we change the bounds to $[0, 12]$ and use the same lengthscale as the RFF variant.

For epistemic index $p_Z(z)$, we pre-sample a buffer of 10,000 particles using a low-discrepancy Sobol sequence with a burn-in of 100, and apply a unit Gaussian icdf to get quasi-random Gaussian samples. During inference or training of all variants, we sample $z$ uniformly from the buffer.

\textbf{MAP Greedy Baseline.} We use the same architecture and training hyperparameters as the base network of the Linear and RFF Epinet variants. We then form our greedy baseline by scoring each compound using this surrogate and selecting the top batch.

\setlength{\tabcolsep}{1.5pt}
\begin{table}[h!]
\small
\centering
\begin{tabular}{|l|c|c|c|}
\hline

\textbf{Baseline} & \makecell{\textbf{Norm. pIC50} \\ \textbf{Iter=50} ($\uparrow$)} & \makecell{\textbf{IC50-Fold} ($\uparrow$)} & \textbf{AUC} ($\uparrow$) \\
\hline
Random & $0.830 \pm 0.003$ & $1.0\times$ & $39.80 \pm 0.13$ \\
\hline
MAP + Greedy & $0.869 \pm 0.007$ & $2.8\times$ & $41.72 \pm 0.32$ \\
\hline
Linear + qPO & $0.890 \pm 0.003$ & $4.9\times$ & $41.56 \pm 0.11$ \\
Linear + EMAX & $0.897 \pm 0.002$ & $5.9\times$ & $42.63 \pm 0.10$ \\
\hline
RFF + qPO & $0.903 \pm 0.001$ & $6.9\times$ & $43.63 \pm 0.10$ \\
RFF + EMAX & $0.904 \pm 0.002$ & $7.1\times$ & $43.74 \pm 0.09$ \\
\hline
Pretrained + qPO & $0.911 \pm 0.002$ & $8.5\times$ & $44.35 \pm 0.10$ \\
Pretrained + EMAX & $0.913 \pm 0.002$ & $9.0\times$ & $44.35 \pm 0.09$ \\
\hline
\end{tabular}
\caption{Comparison of normalized Top-10 mean pIC50 acquired at the final iteration (Iter=50), and area under the optimization curve (AUC) for combinations of Epinets surrogates and acquisition strategies on the EGFR dataset. IC50-Fold indicates the fold improvement in IC50 over Random baseline at the final iteration. The Pretrained + EMAX baseline yields a 3x improvement in final IC50 over the MAP + Greedy baseline and 9x improvement over Random. We report mean and standard errors over 20 random seeds. See corresponding Figure \ref{fig:egfr_scaling}. Higher $\uparrow$ is better.}
\label{tab:egfr_results}
\vspace{.5em}
\end{table}
\setlength{\tabcolsep}{6pt}

\label{sec:EGFR_exp}
\textbf{Running Batch BO.} We aim to maximize experimental binding affinity measured in pIC50 units. At each of the 50 iterations, we retrain the chosen surrogate model from initialization, and acquire a batch of 25 compounds $\{x_1, ..., x_B\}$ in parallel for noiseless evaluation (pre-computed) before adding to the dataset $D_t = \{(x_i, y_i)\}_{i=1}^{Bt}$. For acquisition we use EMAX, qPO, or simply score in a greedy fashion. For MC approximations to the parallel acquisition functions we use 5,000 samples (see Figure \ref{fig:particle_convergence} and \ref{fig:tarray}), and for EMAX we use 10,000 swaps in the inner-loop. In the first iteration we add 100 compounds from the dataset to warmstart training of the surrogate. In order to simulate a realistic setting in which one requires exploration to find the maximizer, these 100 compounds are uniformly sampled after excluding a fraction nearest to the most potent compound by Euclidean distance using COATI embeddings. We provide the code to generate these results at \href{https://github.com/terraytherapeutics/terramax}{\texttt{github.com/terraytherapeutics/terramax}}.

\textbf{EGFR Results}. Figure \ref{fig:egfr_scaling} and Table \ref{tab:egfr_results} summarizes the behavior of each combination of Epinet variant and parallel acquisition function on the EGFR dataset. Many of the Epinet variants that utilize parallel acquisition functions require nearly 5x fewer iterations to obtain the same maximum pIC50 as the greedy baseline, demonstrating that joint samples from an Epinet are highly useful in driving the exploration and hedging required for successful Batch BO. Notably, the combination of the Pretrained Epinet variant and EMAX consistently obtains the most potent compounds at the final iteration for this task. We also note that the three Epinet variants are clearly distinguished at top-10 pIC50 acquisition in early iterations, where there is little data and the prior function is important. These results are promising, however, we acknowledge that these experiments are simplified to a single target and use ligand-only representations. We emphasize that pretraining the prior network on synthetic data would be amenable to representations from more sophisticated models, e.g., by using pair representations from a Pairformer \cite{Abramson:2024vm}.

\subsection{tArray Library Screening}

In this related experiment, we forgo prior network ablations and instead check the importance of being able to rapidly sample large numbers of particles from an Epinet in order to obtain converged estimates of EMAX and qPO, and confirm that this is an important source of efficiency in Batch BO.

\textbf{Dataset.} We collected 50,000 unique and blinded compounds from an internal library of 2.3 million molecules that was screened against a pipeline target. Terray’s Experimentation Meets Machine Intelligence (EMMI) platform drives discovery across a rich pipeline of challenging targets. The experimental side leverages both a proprietary ultradense microarray (tArray) and a highly automated lab. tArray measures millions of interactions between small molecule libraries and targets in a matter of minutes. Target-ligand interaction is quantified via fluorescence intensity for a given molecule relative to the negative control (fold-over, FO).

\begin{figure}[h]
  \centering
  \vspace{0.2em}
  \includegraphics[width=0.475\textwidth]
  {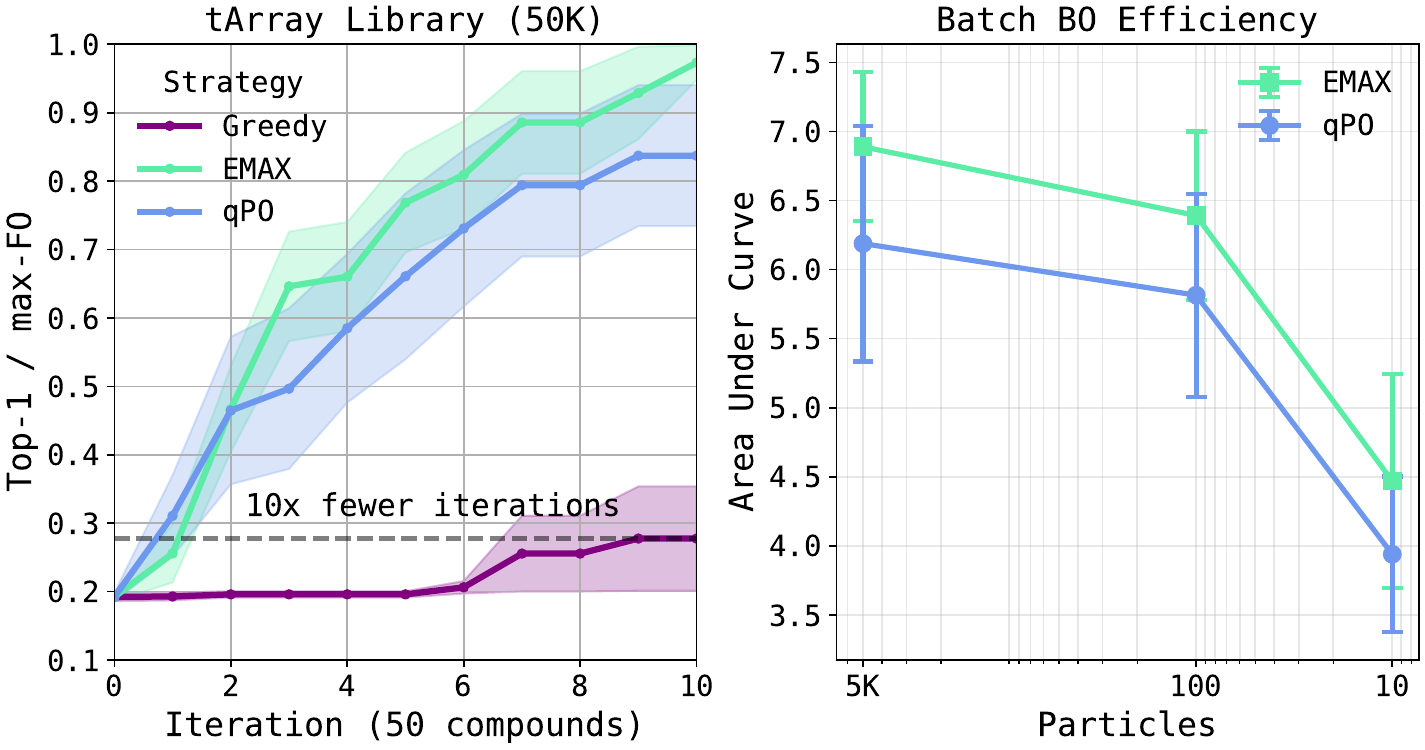}
   \caption{Left: shows normalized Top-1 FO obtained over 10 iterations using 5000 particles with different acquisition strategies, repeated across 10 random seeds. Right: shows degradation of Batch BO efficiency, as measured by area under the normalized Top-1 FO curve, as we decrease the number of particles from 5000 to 10. This shows that the ability to rapidly sample from an Epinet's joint predictive distribution is important for Batch BO efficiency, usually infeasible with more expensive ensemble-based methods.}
   \vspace{0.1em}
  \label{fig:tarray}
\end{figure}

\textbf{Running Batch BO.} We aim to maximize experimental fold-over (FO). We ablate different numbers of particles used for joint samples $[10, 100, 5000]$, where 5000 was the number of samples used in Section \ref{sec:EGFR_exp}. For this dataset we use a Linear Epinet to focus purely on the convergence properties of the procedure. Unless otherwise stated, we follow the same details as the EGFR experiment in Section \ref{sec:EGFR_exp}, except we initialize with 5000 compounds in the first iteration instead of 100, and acquire 50 compounds per iteration, stopping at a total of 10 iterations.

\textbf{tArray Results.} Figure \ref{fig:tarray} shows the efficiency of parallel acquisition functions powered by the joint predictive distribution of an Epinet that uses latent representations from COATI, a chemistry-specific foundation model \cite{kaufman2024coati}. In the first iteration, EMAX and qPO reliably exceed the highest fold-over (FO) obtained by a greedy strategy. Most importantly, as expected, performance does degrade when reducing the number of particles, as measured by area under the Top-1 FO curves, validating that this is required for the overall efficiency of the Batch BO procedure. This is easily accomplished here as obtaining 5000 samples from an Epinet's joint predictive distribution on 50,000 compounds requires 13 seconds on one Nvidia A100 GPU, under a batched forward pass.

\section{Conclusion}
\label{sec:conclusion}

In this work, we presented and validated a scalable surrogate model for Batch BO that demonstrates significant sample-efficiency gains for molecular discovery. An important contribution was to enhance the Epistemic Neural Network (ENN) framework by pretraining the prior network on synthetic data, which we show provides more effective joint predictions when using the right reference process. While we used ligand-only representations for a single target in this work, we emphasize that this framework offers a promising path for accelerating future discovery campaigns by incorporating more complex structure-aware latent representations, other forms of synthetic data-driven priors, as well as extending it to other properties such as ADME and beyond. Our experiments validate that the ability to rapidly sample from an ENN's joint predictive distribution is critical to good performance in Batch BO, which we demonstrate in realistic screening evaluations. In particular, our Pretrained Epinet baseline consistently rediscovered potent EGFR inhibitors in 5x fewer iterations than a greedy baseline and reliably recovered more potent molecules than other baselines at the final iteration.

\bibliography{example_paper}
\bibliographystyle{icml2025}

\end{document}